\documentclass[letterpaper, 10 pt, conference]{ieeeconf}  % Comment this line out if you need a4paper

\IEEEoverridecommandlockouts                              % This command is only needed if 
\usepackage[ruled,vlined]{algorithm2e}
\usepackage{algorithmic}
\usepackage{graphicx}
\usepackage{amsmath, amssymb, amsfonts}
\usepackage{url}
\usepackage{hyperref}
\usepackage{wrapfig}
\usepackage{blindtext}
\usepackage{caption}
\usepackage{subcaption} 
\usepackage{textcomp}
\usepackage{xcolor}
\def\BibTeX{{\rm B\kern-.05em{\sc i\kern-.025em b}\kern-.08em
    T\kern-.1667em\lower.7ex\hbox{E}\kern-.125emX}}
\begin{document}

\title{\bf Continual Model-Based Reinforcement Learning with Hypernetworks}

\author{Yizhou Huang$^{1}$, Kevin Xie$^{2}$, Homanga Bharadhwaj$^{2}$, and Florian Shkurti$^{2}$% <-this % stops a space
%\thanks{*This work was not supported by any organization}% <-this % stops a space
\thanks{$^{1}$Author is with Division of Engineering Science, University of Toronto, Canada. Email: {\tt\small phuang@cs.toronto.edu}}%
\thanks{$^{2}$Authors are with Department of Computer Science, University of Toronto, Canada
        {\tt\small }}%
} 

\newcommand{\Homanga}[1]{\textcolor{red}{Homanga - #1}}
\newcommand{\Florian}[1]{\textcolor{blue}{Florian - #1}}
\newcommand{\Kevin}[1]{\textcolor{green}{Kevin - #1}}
\newcommand{\Philip}[1]{\textcolor{yellow}{Philip - #1}}

\newcommand{\algoName}{\texttt{HyperCRL}}

% \author{
%   Yizhou Huang \quad Kevin Xie \quad Homanga Bharadhwaj \quad Florian Shkurti\\
%   Department of Computer Science\\
%   University of Toronto, Canada\\
%   University of Toronto Robotics Institute\\
%   \texttt{philipyizhou.huang@mail.utoronto.ca}\\ 
%   \{\texttt{kevincxie, homanga, florian}\}\texttt{@cs.toronto.edu} \\
 
% }

\maketitle

%===============================================================================

\begin{abstract}
Effective planning in model-based reinforcement learning (MBRL) and model-predictive control (MPC) relies on the accuracy of the learned dynamics model. In many instances of MBRL and MPC, this model is assumed to be stationary and is periodically re-trained from scratch on state transition experience collected from the beginning of environment interactions. This implies that the time required to train the dynamics model -- and the pause required between plan executions -- grows linearly with the size of the collected experience. We argue that this is too slow for lifelong robot learning and propose \algoName, a method that continually learns the encountered dynamics in a sequence of tasks using task-conditional hypernetworks. 
Our method has three main attributes: first, it includes dynamics learning sessions that do not revisit training data from previous tasks, so it only needs to store the most recent fixed-size portion of the state transition experience; second, it uses fixed-capacity hypernetworks to represent non-stationary and task-aware dynamics; third, it outperforms existing continual learning alternatives that rely on fixed-capacity networks, and does competitively with baselines that remember an ever increasing coreset of past experience. We show that \algoName~is effective in continual model-based reinforcement learning in robot locomotion and manipulation scenarios, such as tasks involving pushing and door opening. Our project website with videos is at this link \url{http://rvl.cs.toronto.edu/blog/hypercrl/}%{http://rvl.cs.toronto.edu/blog/hypercrl/}.
\end{abstract}

%===============================================================================

\section{Introduction}

%
% 1. What's the problem we tackle and what do we propose to solve it?
%
%
% 2. What are some important scenarios and examples of non-stationarity in RL and MPC?
% - changing f(x) for already seen x
% - exploring new x 
%
% - new types of terrains
% - new objects
% - 

Lifelong model-based robot learning is predicated upon continual adaptation to the dynamics of new tasks. For example, robots need to learn to manipulate unseen objects with various mass distributions, walk on new types of terrains with different friction, elasticity, and other physical properties, or even learn to adapt to different tasks, such as walking, running, or climbing stairs. This presents at least two challenges for many model-based reinforcement learning (MBRL) and model-predictive control (MPC) formulations, which typically comprise of a dynamics learning phase followed by a planning/policy optimization and execution phase. First, these methods are not scalable because the time required to train the dynamics model grows linearly with the size of the collected experience. Second, as the robot learner encounters and adapts to new tasks, it has to avoid catastrophic forgetting of the dynamics of old tasks.

In this work, we propose to extend the task-aware continual learning approach based on hypernetworks in \cite{continualhyper} to adapt to changing environment dynamics and to address the scalability and catastrophic forgetting challenges mentioned above in a reinforcement learning setting. We use task-conditional hypernetworks, which are neural network models that accept a learned task encoding as an input, and output the weights of another (target) network. In our case, the output is the dynamics model for that task. No additional information other than task transition boundary is needed. We show that continual learning with hypernetworks leads to effective model-based reinforcement learning, while reducing the number of updates of the dynamics model across planning sessions and preventing catastrophic forgetting.

%
% 3. Why did we adopt the task-aware, fixed-capacity, fixed-dataset-size setting? Why hypernetworks?
%
We consider the setting where task boundaries are known in order to simplify the problem, although there are continual learning methods that have addressed the task-agnostic setting using Bayesian non-parametric methods \cite{xu2020taskagnostic}. In addition, we focus on fixed-capacity hypernetworks and target networks that can handle a sequence of tasks without adding new neurons or layers to the network, unlike many related works \cite{threescenario_cl, pnn} in which every new task adds capacity to the dynamics model. We argue that the fixed-capacity setting, together with storing only the most recent portion of the state transition experience, is more realistic and scalable for lifelong robot learning applications compared to approaches in which the training time or the model's size scales linearly with the size of collected experience.      

%
% 4. What are our main results and contributions?
% 
\begin{figure}[t!]
    \centering
    \vspace{-0.3cm}
    \includegraphics[width=\columnwidth]{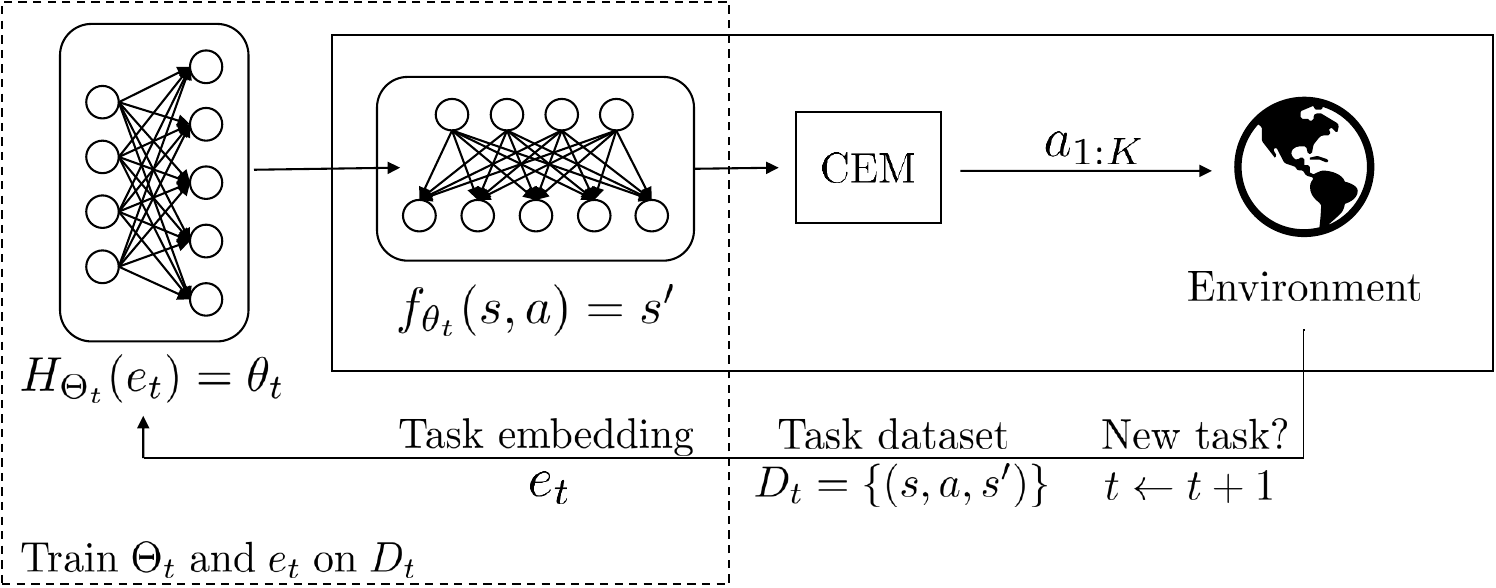}
    \caption{Overview of our proposed method \algoName}
    \vspace{-0.5cm}
    \label{fig:main_diagram}
\end{figure}
Our work makes the following contributions: we show that task-aware continual learning with hypernetworks is an effective and practical way to adapt to new tasks and changing dynamics for model-based reinforcement learning without keeping state transitions from old tasks nor adding capacity to the dynamics model. We evaluate our method on locomotion and manipulation scenarios, where we show that our method outperforms related continual learning baselines.

\section{Related Works}

\textbf{Continual Learning in Neural Networks} Continual learning studies the problem of incrementally learning from a sequential stream of data with only a small portion of the data available at once~\cite{clsurvey}. %Closely related to continual learning, transfer learning refers to using knowledge of one task to assist in a different task. 
A simple yet effective approach is finetuning, which directly tunes the trained source task network on the target task~\cite{bengio2012deep}. 
The efficacy of this approach for continual learning suffers from the well-established phenomenon of catastrophic forgetting~\cite{pnn}. 
Sequential Bayesian posterior updates are a principled way to perform continual learning and naturally avoid the forgetting problem since the exact posterior fully incorporates all previous data but in practice approximations have to be made that may be prone to forgetting.
Elastic Weight Consolidation (EWC) uses a Laplace approximation to the posterior, storing previous tasks' empirical fisher matrices and regularizing future task weight deviations under their induced norms~\cite{ewc}. Other works have also employed variational mean-field approximations~\cite{vcl} or block-diagonal Kronecker factored Laplace approximations~\cite{kfac-cl}.
Synaptic Intelligence (SI) forgoes an obvious approximate Bayesian interpretation but operates similarly to EWC in that it computes a relative parameter importance measure, but through a linear approximation to the contribution in loss reduction due to each parameter on previous tasks~\cite{si}.
%One line of work improves on this by determining the relative importance of different parameters in the network for the previous tasks and proportionally penalizing change in important parameters during transfer. For example, Elastic Weight Consolidation (EWC) uses a quadratic approximation to the loss in the form of the empirical fisher matrix~\cite{ewc}, while Synaptic Intelligence (SI) computes this relative importance through a linear approximation to the contribution in loss reduction due to each parameter on previous tasks~\cite{si}.
%EWC can be interpreted as a Laplace approximation to the principled Bayesian posterior update solution. Variational mean-field approximations have been used in this way as well~\cite{vcl}.
Coreset methods prevent catastrophic forgetting by choosing and storing a significantly smaller subset of the previous task's data, which is used to rehearse the model during or after finetuning~\cite{gem, agem, tinycoreset}. Similarly, the inducing points used in sparse Gaussian Process (GP) formulations, which can be seen as a type of coreset, has been used for continual learning ~\cite{Titsias2020FunctionalRF, vargp}. 
%A variant on coresets are generative replay methods which learn to compress the dataset distribution into a parametric generative model whose samples can be used for retraining.
Another type of approach learns separate task-specific network components. The most common version of this are  multi-head networks that learn and switch between separate output layers depending on the task~\cite{lwf}. Progressive Neural Networks (PNN)~\cite{pnn} are an extreme version of this approach, in which an entirely new copy of the network is appended for each task, thereby eliminating any forgetting. These methods can incur significant memory and compute cost especially for larger models and many tasks.

\textbf{Model-based RL} Model-based reinforcement learning approaches incorporate model learning of the environment in solving the control task.
Traditionally, the model is trained to approximate the stationary dynamics and/or reward of the environment from all collected samples. 
Various choices for the model have been proposed.
Many non-parametric models, such as commonly used GPs, rely on storing and making inferences with past data~\cite{deisenroth2011pilco}, although in many cases the amount of data needed to be stored can be drastically reduced through sparse variational inference~\cite{burt2018explicit}. Nonlinear parametric models typically do not admit efficient sequential update rules and therefore usually train on all past data as well~\cite{raiko2009variational}.
Typically, the trained model is then used to simulate imagined trajectories, either for the purpose of online planning~\cite{pets, planet} or as training data for an amortized policy~\cite{dreamer}.
In non-stationary environments, the dynamics can change over time. 
In this setting, a lot of works focus on quickly adapting to the change in dynamics to minimize online regret, as opposed to retaining performance on previously experienced dynamics~\cite{basso2009reinforcementnonstat,padakandla2019reinforcement, hallak2015contextual}. Meta-learning is a popular tool in this paradigm where a global dynamics model is ``meta-trained" to quickly adapt to the true online dynamics from only a few samples. However, the meta-learning process typically requires updating the meta-model with data from a diverse set of dynamics that is obtained by storing previous experiences in a buffer and/or being able to simultaneously interact with many different environments~\cite{al2017continuous}.

\textbf{Relationship to Meta-Learning} Our work is different from meta-learning approaches in MBRL like \cite{mole_mbrl, learningtoadapt} in two ways. First, we focus on preventing catastrophic forgetting and do not explicitly train a model prior across multiple tasks for fast adaptation. This means that we do not need to design a set of tasks for meta-training and in principle, our work can continuously learn to perform new tasks from scratch. Second, we do not require the use of a replay buffer that grows linearly with the number of tasks or total length of collected state-transition pairs. We emphasize that this is consistent with the theme of continual learning, where storing past data is limited.

\textbf{Continual RL} Memory-efficient continual learning methods in the reinforcement learning setting have also been proposed. PNN was used in an on-policy actor-critic method and was demonstrated on sequential discrete action Atari games. The authors of~\cite{p&c} build on top of PNN and continual policy compression methods~\cite{Berseth2018ProgressiveRL} by compressing the extended model from PNN into a fixed size network after each task. For the compression stage, they propose a more scalable online EWC algorithm that eschews the linear cost of storing past fisher matrices. 
The use of coresets has also been explored in this setting~\cite{abel2018state}. In~\cite{mole_mbrl}, a mixture model of separate task-specific neural networks is used for the environment model that requires adding a new model each time a task is instantiated.
%Additionally, although they test their method on simulated continuous control locomotion tasks, the performance of their method crucially relies upon a meta-learning pretraining stage that violates the continual learning constraints. Their baselines without this component perform much worse.
A similar approach was also demonstrated using a mixture of GPs using an online clustering algorithm~\cite{abdollahi2015adaptive}. Our work is also related to MPC interpretations as a reduction to online learning~\cite{boots_mpc_online_learning}.

\textbf{Robust and Adaptive Control} Existing literature on control theory provides many related classes of approaches that handle changes in dynamics: adaptive control methods handle unknown parameters of a dynamics model by estimating them over time so as to execute a given trajectory, and robust control methods provide stability guarantees as long as the parameter or model disturbance is within bounds \cite{adaptivecontrol_book}. The setting we study in this work differs in that we learn the dynamics model from scratch, without assuming a reference model, and it may change over tasks without imposing any bounds on particular parameters.   

\section{Preliminaries}
\textbf{Hypernetworks for Continual Learning.} A hypernetwork~\cite{hypernetworks,Schmidhuber1991LearningTC} is a network that generates the weights of another neural network. %The hypernetwork takes a set of inputs that contain some information about the structure of the weights (for example an embedding for each layer's description) and generates the weights for each layer. 
The hypernetwork $H_{\Theta}(e)=\theta$ with weights $\Theta$ can be conditioned on an embedding vector $e$ to output the weights $\theta$ of the main (target) network $f_{\theta}(x) = f(x;\theta) = f(x;H_{\Theta}(e))$ by varying the embedding vector $e$. The hypernetwork is typically larger with respect to the number of trainable parameters in comparison to the main network because the size of output layer in the hypernetwork is equal to the number of weights in the target network. Hypernetworks have been shown to be useful in the continual learning setting~\cite{continualhyper} for classification and generative models. %, where instead of trying to adapt the weights of the main network across $T$ tasks, a task-conditioned hypernetwork is learned instead to predict the weights of the main network for each successive task. 
This has been shown to alleviate some of the issues of \textit{catastrophic forgetting}. They have also been used to enable gradient-based hyperparameter optimization~\cite{stn_vicol}. %that occurs when a network is directly trained on a sequence of tasks without retaining information about past tasks. %The parameters of the hypernetwork are updated by backpropagating gradients from all the task losses. 

% \textbf{Planning with CEM and MPC} The Cross-Entropy Method (CEM)~\cite{cem} is a widely used online planning algorithm, when a dynamics model and optionally a reward model of the environment is learned. Model-Predictive Control (MPC) searches for an optimal sequence of actions under the learned model and executes the first action of the sequence in the environment, and re-plans. Executing actions via MPC and iterative re-planning helps alleviate issues of compounding errors in planning~\cite{vimpc}. %CEM is a random shooting based search heuristic to perform this planning of action sequences. 
% CEM samples action sequences from a time-evolving distribution which is usually considered to be a diagonal Gaussian $a_{1:h}\sim\mathcal{N}(\mu_{1:h},\texttt{diag}(\sigma^2_{1:h}))$,  where $h$ is the planning horizon. Action sequences are iteratively re-sampled, evaluated under the currently learned dynamics model, and the sampling distribution parameters $\mu_{1:h},\sigma_{1:h}$ are re-fitted to the top percentile of trajectories.  CEM for planning in MBRL has been successfully used in a number of previous approaches~\cite{pets,planet}, as it alleviates exploitation of model-bias compared to purely gradient based optimizations~\cite{l4dc}, and can better adapt to varying dynamics  as compared to fully amortized policies~\cite{dreamer}.  

\textbf{Planning with CEM and MPC.}
%The Cross-Entropy Method (CEM)~\cite{cem} is a widely used online planning algorithm, when a dynamics model and optionally a reward model of the environment is learned. Model-Predictive Control (MPC) searches for an optimal sequence of actions under the learned model and executes the first action of the sequence in the environment, and re-plans. Executing actions via MPC and iterative re-planning helps alleviate issues of compounding errors in planning~\cite{vimpc}. %CEM is a random shooting based search heuristic to perform this planning of action sequences. 
The Cross-Entropy Method (CEM)~\cite{cem} is a widely used online planning algorithm that samples action sequences from a time-evolving distribution which is usually considered to be a diagonal Gaussian $a_{1:h}\sim\mathcal{N}(\mu_{1:h},\texttt{diag}(\sigma^2_{1:h}))$,  where $h$ is the planning horizon. Action sequences are iteratively re-sampled and evaluated under the currently learned dynamics model, and the sampling distribution parameters $\mu_{1:h},\sigma_{1:h}$ are re-fitted to the top percentile of trajectories. CEM for planning in MBRL has been successfully used in a number of previous approaches~\cite{pets,planet}, as it alleviates exploitation of model bias compared to purely gradient based optimizations~\cite{l4dc} and can better adapt to varying dynamics  as compared to fully amortized policies~\cite{dreamer}.

\section{The Proposed Approach}
%We describe the overall continual learning algorithm for adapting to different tasks online, and discuss the specifics of each component in detail.
\subsection{Problem Setting and Method Overview}
We consider the following problem setting: A robot interacts with the environment to solve a sequence of $T$ goal-directed tasks, each of which brings about different dynamics while having 
%the same goal (accesed via a task-independent reward function $r(s)$), 
the same state-space $\mathcal{S}$ and action space $\mathcal{A}$. The robot is exposed to the tasks sequentially online without revisiting data collected in a previous task.
The robot also has finite memory and is not allowed to maintain a full history of state transitions for the purpose of re-training. Since the distribution of tasks changes over time and the agent does not know about it a priori, it must continually adapt to the streaming data of observations that it encounters, while trying to solve each task. The robot knows when a task switch occurs.

% This problem setting is general and encompasses a wide range of real-world robotic control problems e.g. unknown and temporally varying system parameters, motor and joint malfunctions on the robot, and varying local map of the environment in navigation. 

% Formally, we consider a set of $T$ sequential tasks, an underlying task-dependent transition dynamics $p(s'|s,a,t)$ ($t\in[1,..,T]$), and an underlying task-independent reward function $r(s)$ that depends on the goal being solved. Both the transition dynamics and reward function are unknown to the agent. Let $\mathcal{S}$ denote the state-space exposed to the agent, and $\mathcal{A}$ denote the action space the agent can control, both of which are task-independent. Let $K$ be the overall length of the planning horizon. The objective is to learn a planning algorithm that maximizes cumulative sum of discounted rewards across $T$ tasks. 

\begin{algorithm}[]
    \caption{Continual Reinforcement Learning via Hypernetworks (\algoName)}
    \label{alg:lilac}
    \begin{algorithmic}[1]
    \STATE {\bfseries Input:} $T$ tasks, each with its own dynamics $\mathcal{S} \times \mathcal{A} \rightarrow \mathcal{S'}$, reward $r(s,a)$. Learning rates $\alpha_\Theta$, $\alpha_e$, and planning horizon $h$. \\
    \STATE Randomly initialize hypernetwork weights $\Theta_1$
    \FOR {task $t$ = 1, 2, $\dots$, $T$}
        \STATE Initialize task-specific replay buffer $\mathcal{D}_t = \{\}$
        \STATE Randomly initialize task embedding $e_{t}$
        \STATE Collect $P$ episodes of trajectories $\tau$ using a random policy and add it to $\mathcal{D}_t$
        \FOR {episode $m$ = 1, 2, \dots, $M$}
            \STATE (Optionally) Reset the environment; Observe $s_0$ %\Homanga{We should clarify that the hard reset step is not strictly needed? for example when deploying on a real robot} \Philip{Yes, the algorithm doesn't need to be episodic}
            \STATE Generate target network weights $\theta_t = H_{\Theta_t}(e_{t})$
            \STATE // \texttt{The dynamics model for the current episode is $f_{\theta_t}(\cdot)$}
            \FOR {step $k$ = 1, 2, \dots, $K$}
                \STATE Optimize action sequence $a_{k:k+h}$ using CEM with $f_{\theta_t}(\cdot)$ and known reward 
                \STATE Execute first action $a_k$, observe next state $ s_{k+1}$, add $(s_k, a_k, s_{k+1})$ to $\mathcal{D}_t$ 
               % \STATE $ \mathcal{D} \gets \mathcal{D} \cup (s_k, a_k, s_{k+1}) $ 
            \ENDFOR
            \STATE // \texttt{Update hypernetwork and task embedding}
            \FOR {$s$ = 1, 2, \dots, $S$}
                \STATE Sample a batch $\mathcal{B}$ of state-action pairs $(s_k, a_k, s_{k+1})$ from $\mathcal{D}_t$ and compute $\mathcal{L}_t$
                                \STATE $\Theta_t \gets \Theta_t - \alpha_\Theta \nabla_{\Theta_t} \mathcal{L}_t(\Theta_t, e_t)$
                \STATE $e_t \gets e_t - \alpha_e \nabla_{e_t} \mathcal{L}_t(\Theta_t, e_t)$ 
            \ENDFOR
        \ENDFOR
        \STATE $\Theta_{t+1} = \Theta_{t}$
    \ENDFOR
    \end{algorithmic}
   % \vspace*{-0.5cm}
\end{algorithm}
%\vspace*{-0.9cm}

We consider the solution setting of MBRL with a learned dynamics model, the parameters of which are inferred through a task-conditioned hypernetwork. Given learned task embeddings $e_t$ and parameters $\Theta_t$ of the hypernetwork $H(\cdot)$, we infer parameters $\theta_{t}$ of the dynamics neural network $f_{\theta_t}(\cdot)$. Using this dynamics model, we perform CEM optimization to generate action sequences and execute them in the environment for $K$ time-steps with MPC. We store the observed transitions in the replay dataset and update the parameters of the hypernetwork $\Theta_t$ and task-embeddings $e_t$ (off-policy optimization). We repeat this for $M$ episodes per task, and for each of the $T$ tasks sequentially.
\subsection{Training Procedure}

\noindent\textbf{Dynamics Learning.} The learned dynamics model is a feed-forward neural network whose parameters vary across tasks. One way to learn a dynamics network $f_\theta(\cdot)$ across tasks is to update it sequentially as training progresses. However, since our problem setting is such that the agent is not allowed to retain state-transition data from previous tasks in the replay buffer, adapting the weights of a single network sequentially across tasks is likely to lead to catastrophic forgetting~\cite{continualhyper}. In order to alleviate issues of catastrophic forgetting while trying to adapt the weights of the network, we learn a hypernetwork that takes task embeddings as input, and outputs parameters for the dynamics network corresponding to every task, learning different dynamics networks $f_{\theta_t}(\cdot)$ for each task $t$.

We assume that the agent has finite memory and does not have access to state-transition data across tasks. So, the task-specific replay buffer $\mathcal{D}_t$ is reset at the start of every task $t$. For the current episode, the agent generates a dynamics network $f_{\theta_t}$ using $\theta_t=H_{\Theta_t}(e_t)$. Then, for $k=1...K$ timesteps and planning horizon $h$, the agent optimizes action sequences $a_{k:k+h}$ using CEM, and executes the first action $a_k$ (MPC).  $\mathcal{D}_t$ is augmented by a tuple $(s_k,a_k,s_{k+1})$, where $s_k$ is the current state, $a_k$ is the executed action, and $s_{k+1}$ is the next observed state under task $t$. 

The parameters $\Theta_t$ of the hypernetwork and the task embeddings $e_t$ are updated by backpropagating gradients with respect to the sum of a dynamics loss $\mathcal{L}_\text{dyn}$ and a regularization term. We define the dynamics loss as $\mathcal{L}_{dyn}(\Theta_t, e_t)=\sum_{\mathcal{D}_t}||\hat{s}_{k+1}-s_{k+1} ||_2$, where the predicted next states are $\hat{s}_{k+1}=f_{\theta_t}(s_k,a_k)$ and $\theta_t=H_{\Theta_t}(e_t)$.
%$\mathcal{L}_\text{dyn}$ is computed by sampling a batch of task-specific transitions $\{(s_k,a_k,s_{k+1})\}$ from $\mathcal{D}_t$, inferring the predicted $\hat{s}_{k+1}=f_{\theta_t}(s_k,a_k)$ and computing the  sum of $L_2$ distances $\sum||\hat{s}_{k+1}-s_{k+1} ||_2$ across the batch. 
In practice, we infer the difference $\Delta_{k+1}$ through the dynamics network ($\Delta_{k+1}=f_{\theta_t}(s_k,a_k)$) such that $\hat{s}_{k+1} = s_k + \Delta_{k+1}$ for stable training. %\Philip{Actually, some cases (in door and cheetah) I actually use a probabilistic network for dynamics. During CEM sampling I always choose the mean prediction}
In addition, inputs to the $f_{\theta_t}$ network are normalized, following the procedure in previous works~\cite{pets}. Similarly, a new $e_t$ is initialized at the start of every task and updated every episode during the task by gradient descent. Older task embeddings ($e_{1:t-1}$) could also be amenable to gradient descent, but we keep them fixed for simplicity.
\vspace{0.1cm}

\noindent\textbf{Regularizing the Hypernetwork.} To alleviate catastrophic forgetting, we regularize the output of the hypernetwork for all previous task embeddings $e_{1:t-1} $. After training for task $t-1$,  a snapshot of the hypernetwork weights is saved as $\Theta_{t-1}$. For each of the past tasks $i=1...t-1$, we use a regularization loss to keep the outputs of the snapshot $H_{\Theta_{t-1}}(e_i)$ close to the current output $H_{\Theta_{t}}(e_i)$. This approach sidesteps the need to store all past data across tasks, preserves the predictive performance of dynamics networks $f_{\theta_t}$, and only requires a single point in the weight space (a copy of the hypernetwork) to be stored. The task embeddings are differentiable vectors learned along with parameters of the hypernetwork. The overall loss function for updating $\Theta_t$ and $e_t$ is given by the sum of the dynamics loss $\mathcal{L}_\text{dyn}(\cdot)$, which is evaluated on the data collected from task $t$ and the regularization term $\mathcal{L}_\text{reg}(\cdot)$:
%\vspace*{-0.2cm}
\begin{eqnarray}
    \begin{aligned}
      \mathcal{L}_t(\Theta_t, e_t) = \mathcal{L}_\text{dyn} (\Theta_t, e_t) + \mathcal{L}_\text{reg}(\Theta_{t-1},\Theta_t, e_{1:t-1}) \\
      \mathcal{L}_{\text{reg}}(\cdot)
      = \frac{\beta_\text{reg}}{t-1} \sum_{i=1}^{t-1} \| H_{\Theta_{t-1}}(e_i) - H_{\Theta_t}(e_i)\|_2^2
    \end{aligned} 
\end{eqnarray}
The planning objective for CEM optimization of action sequences is given by the sum of rewards obtained by executing the sequence of actions $a_{k:k+h}$ under the learned dynamics model $f_{\theta_t}(\cdot)$ for the task. The reward function $r(s, a)$ is assumed to be known, but nothing precludes learning it from data under our current framework.
%:
%\vspace*{-0.2cm}
%\begin{align*}
%    \mathcal{L}_\text{plan} =  \sum_{g=k,\hat{s}_{g+1}=f_{\theta_t}(\hat{s}_g,a_g)}^{k+h}r(\hat{s}_g,a_g)
%\end{align*}
%\vspace*{-0.5cm}

\vspace{-0.05cm}
\section{Experiments}
We perform multiple robot simulation experiments
%\footnote{Due to Covid-19, we were unable to perform real robot experiments. We have included a 1-page document (as instructed in the submission instructions) about how we can conveniently scale our approach to a real robot system.}
to answer the following questions: (a) How does \algoName~compare with existing continual learning baselines in terms of overall performance across tasks? (b) How effective is \algoName~in preventing catastrophic forgetting across tasks?

% \begin{figure} [hb!]
%     {\includegraphics[width=6cm]{figure/pusher_env.pdf}}
%     {\caption{\textbf{Pusher Environment}. 
%     \Kevin{Ideally the two pictures should be taken from the same camera perspective if possible}\Philip{Done}
%     The blue and green parts of the block have equal size but different densities. The goal is to push the block to the goal indicated by the red dot. \Homanga{This looks good. The plots for Door env can probably go alongside this on the right} \Philip{I was thinking I should put the table below on the right} \Homanga{The table below should probably go in the appendix - we can forward reference it here}}}
%     \label{fig:pusher}
% \end{figure}

\subsection{Pushing a block of non-uniform mass}
\begin{figure} [t!]
     %\vspace{-0.3cm} 
     \begin{subfigure}[b]{0.55\columnwidth}
         \centering
         \includegraphics[width=4cm]{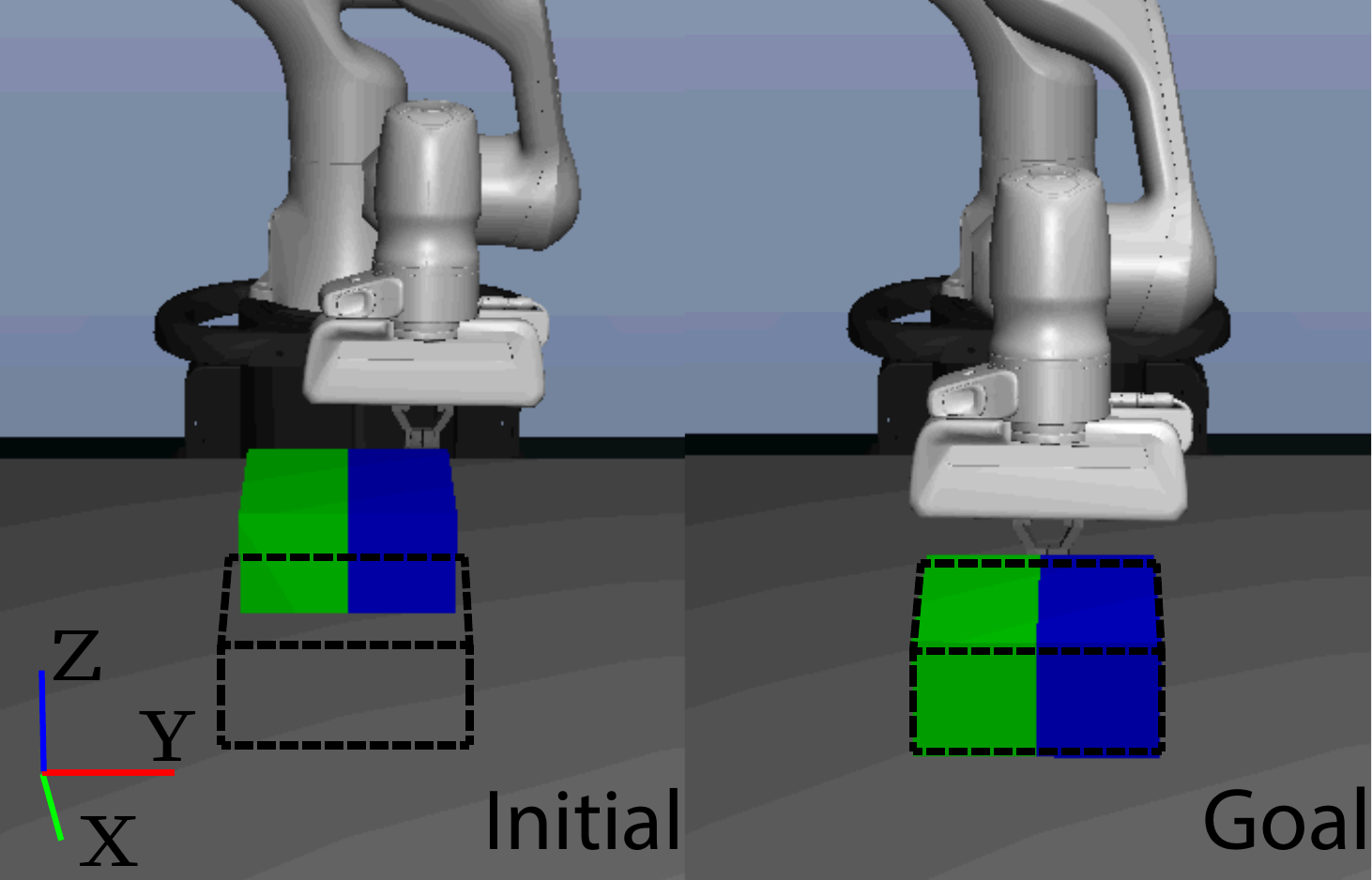}
         \caption{Initial and goal poses of the block.}
        \label{fig:pusher_setting}
     \end{subfigure} 
     \begin{subfigure}[b]{0.42\columnwidth}
         \begin{tabular} {p{0.5cm}  p{0.7cm} p{0.7cm} }
            \hline
            \textbf{Task} & \textbf{Green} & \textbf{Blue} \\
            \hline
            1 & 500 & 500 \\
            \hline
            2 & 100 & 500 \\
            \hline
            3 & 500 & 100 \\
            \hline
            4 & 500 & 250 \\
            \hline
            5 & 250 & 500 \\
            \hline
            \end{tabular}
         \label{table:pusher}
         \caption{Density of the block in $kg/m^3$}
         \label{fig:three sin x}
     \end{subfigure}
    \caption{\textbf{Setup of Pusher Environment}. The robot needs to solve 5 pushing tasks sequentially, each involving a block of different mass distribution. The objective is to push the block to the goal pose (indicated by the dotted frame) without changing its orientation.}
    \label{fig:pusher_illustration}
    \vspace{-0.6cm}
\end{figure}

First, we look at an intuitively simple experiment simulated using Surreal Robotics Suite \cite{corl2018surreal}, in which a Panda robot tries to push a non-uniform-density block across the table (Figure \ref{fig:pusher_illustration}). The objective is to push the block to the goal position while \textit{maintaining its initial orientation}. We vary the densities of the left and right parts of the block across different tasks ($T=5$), changing the center of mass and moment of inertia. The robot needs to learn to maneuver the position of its end-effector on the side of the block to correct for orientation deviations while pushing the block forward. Each episode is 100-step long, which is 10 seconds of simulator time. At the start of each episode, we initialize the robot end-effector to a fixed position behind the block.

The state is represented as a concatenated vector $(x_\text{ee}, x_1,x_2, x_3, x_4)$.
Here $(x_1,x_2, x_3, x_4)$ represents the $xy$ positions of the four corners and $x_\text{ee}$ is the $xy$ position of the end-effector with respect to the block. 
The input action vector, $\delta x_\text{ee}$, specifies a desired offset with respect to the current position $x_\text{ee}$. The robot is actuated using an operational space controller \cite{khatib1987unified}. It calculates the joint torques given an action that is updated at 10Hz. The reward function
is set to minimize the sum of distances between the current and goal pose of the block following $r(s, a) = \sum_{i=1}^4 ( 1 - \text{tanh}(10~||x_i - g_i||)) - 0.25||a||$, where ${g_i}$ is the goal position of the $i^{th}$ corner. The reward function consists of one term per corner within the range [0, 1], and to maximize reward the robot should minimize the corner distances to the goal pose.

\begin{figure}[h!]
   \centering
   \hspace*{-0.3cm}
      \includegraphics[width=0.9\columnwidth]{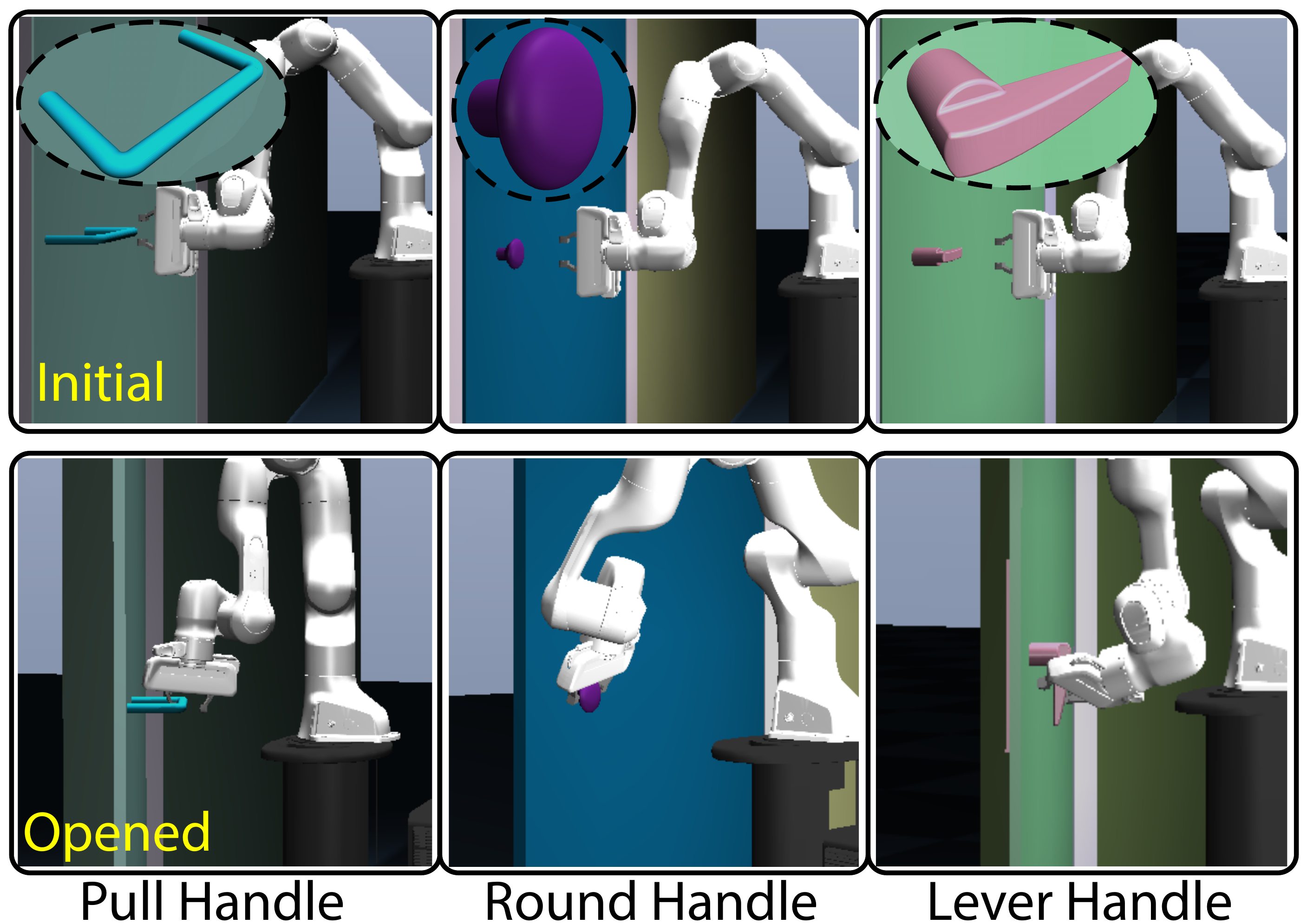} 
      \caption{\textbf{Setup of Door Environment}. For round and lever handles, the robot must first rotate the handle \textit{clockwise} before pulling the door open. Additionally, we introduce two more tasks by flipping the rotational direction to \textit{counter-clockwise} for the round and lever handles.}
    \label{fig:panda}
      \vspace*{-0.3cm}
 \end{figure}

\begin{figure*} [h]
    \centering
    \includegraphics[width=\linewidth]{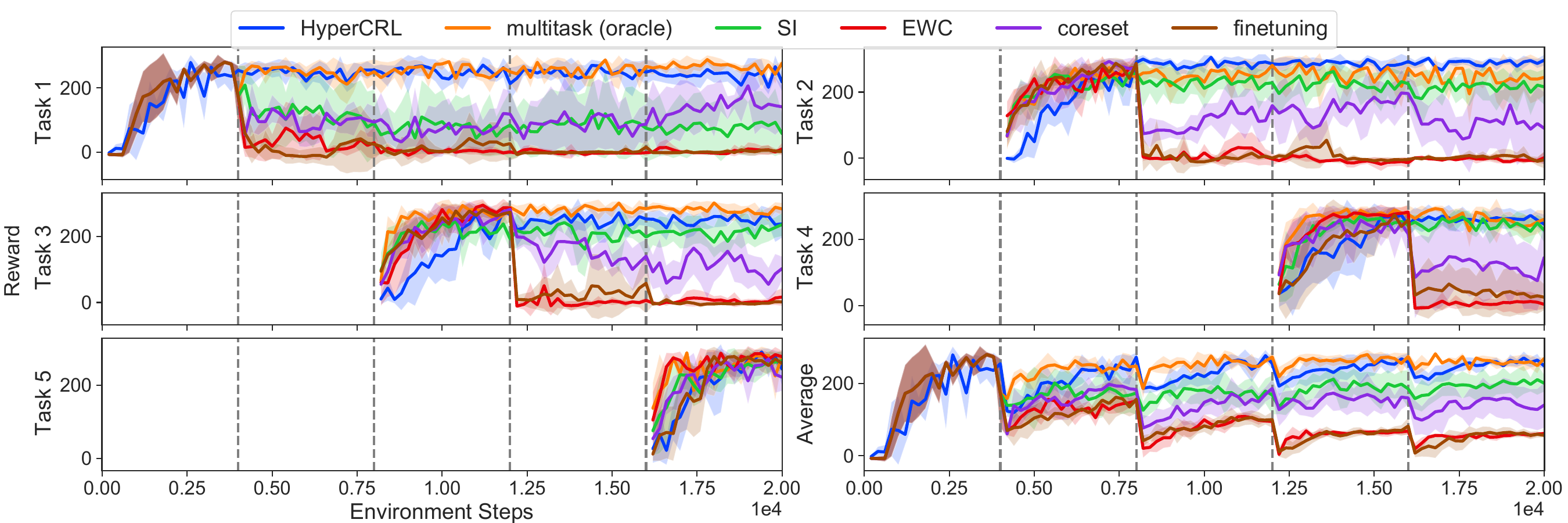}
    \caption{\textbf{Reward on Pusher Environment}. Shown are episodic rewards evaluated during training. Results are averaged across four random seeds, and the shaded areas represent one standard deviation. Each task is trained for 4k steps, summing to 20k steps in total. Vertical dotted lines indicate task switches. The bottom-right subplot shows the average reward across all task $\leq t$ seen so far.}
    \label{fig:pusher_reward}
    \vspace{-0.3cm}
\end{figure*}

\vspace*{-0.2cm}
\subsection{Door Opening}
Next, we experiment on a more complex task to better demonstrate the flexibility of \algoName. We choose a door opening experiment, where the Panda robot has to open doors with different types of handles (Figure \ref{fig:panda}). Unlike the pushing example, the dynamic switches between tasks are much more discontinuous and the tasks require different motions to solve, due to the rotational joints imposed by some of the handles.  A total of $T = 5$ tasks need to be solved sequentially, each involving a different handle type. Tasks 2 and 4 (similarily tasks 3 and 5) both have a round (lever) handle, but with different turning directions. The environment is modified from the DoorGym environment \cite{doorgym2019} and simulated in the Surreal Robotics Suite~\cite{robosuite}. 
 
We model the environment as follow: (i) $\phi_d$ represents the angle of the door (ii) $\phi_k$ represents the angle of the door handle (iii) $x \in \mathbb{R}^3$ is the position, $q \in \mathbb{R}^4 $ is the quaternion representation of the rotation of the robot end-effector with respect to the handle (iv) $d \in \mathbb{R}$ is the state of the gripper (v) $q_j \in \mathbb{R}^7$ is the angle of the joints of the Panda arm. Concatenating all the above elements and their time derivative gives the full state vector $(\phi_d, \dot{\phi}_d, \phi_k, \dot{\phi}_k, x, q, d, q_j, \dot{q}_j) \in \mathbb{R}^{26}$. 
Similar to the pushing environment, the robot is actuated with operational space control updated at 10Hz. The input action is a vector $(\delta x, \delta q, \delta d)$, which specifies a translation and rotational movement of the end-effector in the world frame. Finally, the reward function has five components as follows: $r(s, a) = -||x||_2 - \text{log}(||x||_2 + \epsilon) - ||q_{o}||_2 + 50\,\phi_d + 20\,\phi_k$, where $q_{o}$ is the difference between the orientation of the current end-effector and the required pose for opening the door.
 
\begin{figure}[h!]
   \centering
   \vspace*{-0.3cm}
      \begin{minipage}[c]{0.64\columnwidth}
        \includegraphics[width=0.95\columnwidth]{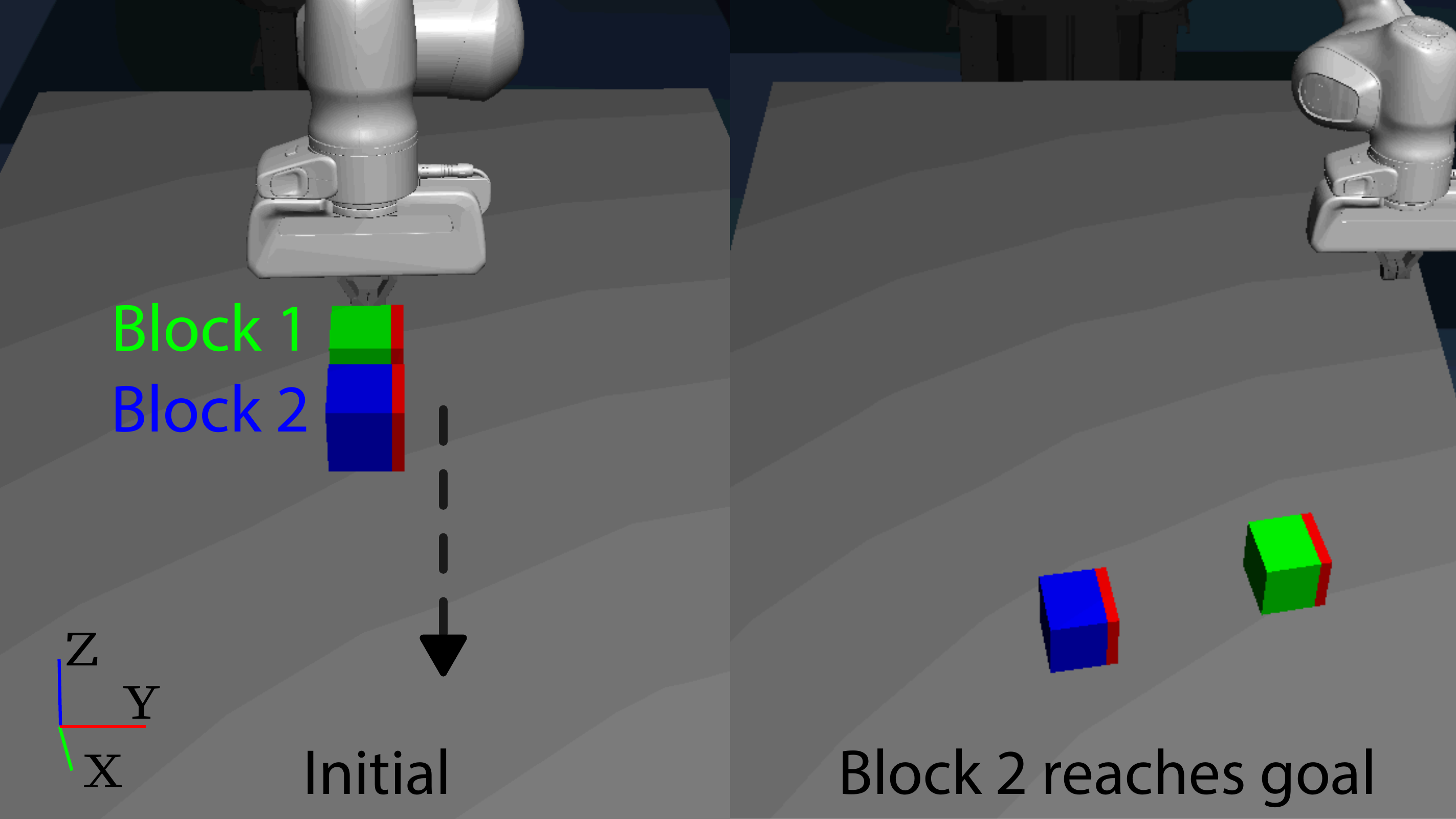} 
      \end{minipage}
      \begin{minipage}[c]{0.3\columnwidth}
         \caption{\textbf{Setup of Slide Environment}. The Panda arm should push block 1, which will kick block 2 in motion towards the goal pose. There are 5 tasks sequentially with different friction for block 2.}
         \label{fig:slide}
      \end{minipage}
     \vspace*{-0.6cm}
  \end{figure}
\vspace{-0.1cm}

\subsection{Block Sliding}
Our last experiment involves a Panda robot and two separate blocks placed on tabletop with low friction. The initial pose of the end-effector, the two blocks are placed along a straight line and we number the two blocks shown in Figure \ref{fig:slide}. To move block 2 to its goal pose, the manipulator should first hit block 1, which would slide away and in turn set the second block into sliding motion until stopped by friction. We vary the friction of block 2 across different tasks ($T$ = 5), while keeping friction of block 1 the same. Similar to the pusher experiment, we represent the state as a concatenated vector $(x_\text{ee}, x^1_{1:4}, x^2_{1:4})$, where $x_\text{ee}$ denotes the $xy$ position of the end-effector, and $(x^1_{1:4}, x^2_{1:4})$ denotes the $xy$ positions of the corners of both blocks. The reward function minimizes the distances between the current and goal pose of the second block $g_{1:4}$ as follows: $r(s, a) = \sum_{i=1}^4 (1 - \text{tanh}(10 ||x^2_i - g_i ||)) - 0.1||a||$. The robot must adjust the momentum of the first block in the initial push for different frictions to successfully deliver the second block to its goal. Each episode is 30-step long and equates 3 seconds in the simulator.
\vspace{-0.1cm}
\subsection{Baselines}
We compare the hypernetwork to the following baselines: (i) Multi-task learning with access to a buffer of all data from all previous tasks (an oracle) (ii) Coreset that remembers one percent of the state-action transition data per task, sampled randomly (iii) Synaptic Intelligence \cite{si} (iv) Elastic Weight Consolidation \cite{ewc} (v) Fine-tuning, where we optimize $f_{\theta_t}(\cdot)$ on each task's data without regularization. All the baseline models resemble the target model in our hypernetwork setup, except the multi-head output layer with one head per task. For coreset or multi-task learning, an additional batch of past data is sampled from the coreset or oracle every update step, and contributes to the total dynamics loss.

\setlength{\tabcolsep}{1pt}

\begin{table} [b!]
\vspace{-15pt}
\begin{tabular}{ p{1.4cm}p{1.4cm}p{1.4cm}p{1.4cm} p{1.4cm} p{1.4cm}}
 \hline
 \multicolumn{6}{c}{ \textbf{\% Retention in terms of Reward (Pusher)}} \\
 \hline
 \textbf{Task} & \textbf{1} & \textbf{2} & \textbf{3} & \textbf{4} & \textbf{Average} \\
\hline
 Multi-task & \textbf{142 $\pm$ 74} & 91 $\pm$ 25 & 96 $\pm$ 9 & 91 $\pm$ 11 & \textbf{106 $\pm$ 20}\\
 HyperCRL  & 99 $\pm$ 10  & \textbf{107 $\pm$ 9} & \textbf{98 $\pm$ 13} &\textbf{103 $\pm$ 15} & 102 $\pm$ 6 \\
 SI  & 40 $\pm$ 54 & 88 $\pm$ 21 & 95 $\pm$ 21 & 92 $\pm$ 14 & 79 $\pm$ 16 \\
 EWC  & 7 $\pm$ 15 & 13 $\pm$ 8 & 6 $\pm$ 8 & 0 $\pm$ 3 & 4 $\pm$ 5 \\
 Coreset & 87 $\pm$ 66 & 32 $\pm$ 46 & 39 $\pm$ 13 & 61 $\pm$ 43 & 55 $\pm$ 23 \\
 Finetuning & 0 $\pm$ 2 & -2 $\pm$ 6 &  1 $\pm$ 3  & 13 $\pm$ 16 & 3 $\pm$ 4\\
 \hline
 \hline
 \multicolumn{6}{c}{ \textbf{\% Retention in terms of Normalized Reward (Door, see Figure \ref{fig:door_reward})}} \\
 \hline
 \textbf{Task} & \textbf{1} & \textbf{2} & \textbf{3} & \textbf{4} & \textbf{Average} \\
 \hline
 Multi-task & 37 $\pm$ 70 & 81 $\pm$ 88 & 59 $\pm$ 60 & \textbf{93 $\pm$ 129} & 68 $\pm$ 45\\
 HyperCRL  & \textbf{113 $\pm$ 75}  & \textbf{100 $\pm$ 76} & \textbf{99 $\pm$ 36} & 86 $\pm$ 68 & \textbf{100 $\pm$ 26} \\
 SI  & -17 $\pm$ 27 & 11 $\pm$ 61 & 6 $\pm$ 37 & 16 $\pm$ 60 & 4 $\pm$ 24 \\ 
 EWC  & -6 $\pm$ 33 & 17 $\pm$ 59 & 4 $\pm$ 25  & 16 $\pm$ 50 & 7 $\pm$ 22 \\
 Coreset & -5 $\pm$ 28 & 26 $\pm$ 53 & 21 $\pm$ 34 & 45 $\pm$ 68 & 22 $\pm$ 24 \\
 Finetuning & -14 $\pm$ 28 & 11 $\pm$ 58 & 3 $\pm$ 23 & 16 $\pm$ 65 & 4 $\pm$ 23 \\
 \hline
 \hline
 \multicolumn{6}{c}{ \textbf{\% Retention in terms of Reward (Block Sliding)}} \\
 \hline
 \textbf{Task} & \textbf{1} & \textbf{2} & \textbf{3} & \textbf{4} & \textbf{Average} \\
 \hline
 Multi-task & \textbf{106 $\pm$ 36} & \textbf{131 $\pm$ 27} & 103 $\pm$ 39 & \textbf{113 $\pm$ 12} & \textbf{113 $\pm$ 16}\\
 HyperCRL  & 82 $\pm$ 56  & 98 $\pm$ 11 & \textbf{109 $\pm$ 16} & 95 $\pm$ 65 & 96 $\pm$ 22 \\
 SI  & 37 $\pm$ 14 & 92 $\pm$ 12 & 68 $\pm$ 56 & 103 $\pm$ 7 & 75 $\pm$ 15 \\ 
 EWC  & -5 $\pm$ 7 & 14 $\pm$ 20 & 13 $\pm$ 35  & 39 $\pm$ 27 & 15 $\pm$ 12 \\
 Coreset & 58 $\pm$ 18 & 97 $\pm$ 72 & 65 $\pm$ 33 & 73 $\pm$ 21 & 73 $\pm$ 21 \\
 Finetuning & -2 $\pm$ 8 & -12 $\pm$ 15 & -1 $\pm$ 4 & -7 $\pm$ 5 & -6 $\pm$ 5 \\
 \hline
\end{tabular}
\caption{\textbf{Performance Retention on Pusher, Door and Block Sliding Envs}. We measure average episodic reward retained at the end of training all 5 tasks compared to the end of training on task $t$. Results show the mean and one std. deviation, evaluated across 4 seeds and 10 episodes per seed. Greater than 100 indicates positive backward transfer, otherwise it indicates forgetting.}
\label{table:pusher_forgetting}
\end{table}

\begin{figure*}[t]
    \centering
    \includegraphics[width=\linewidth]{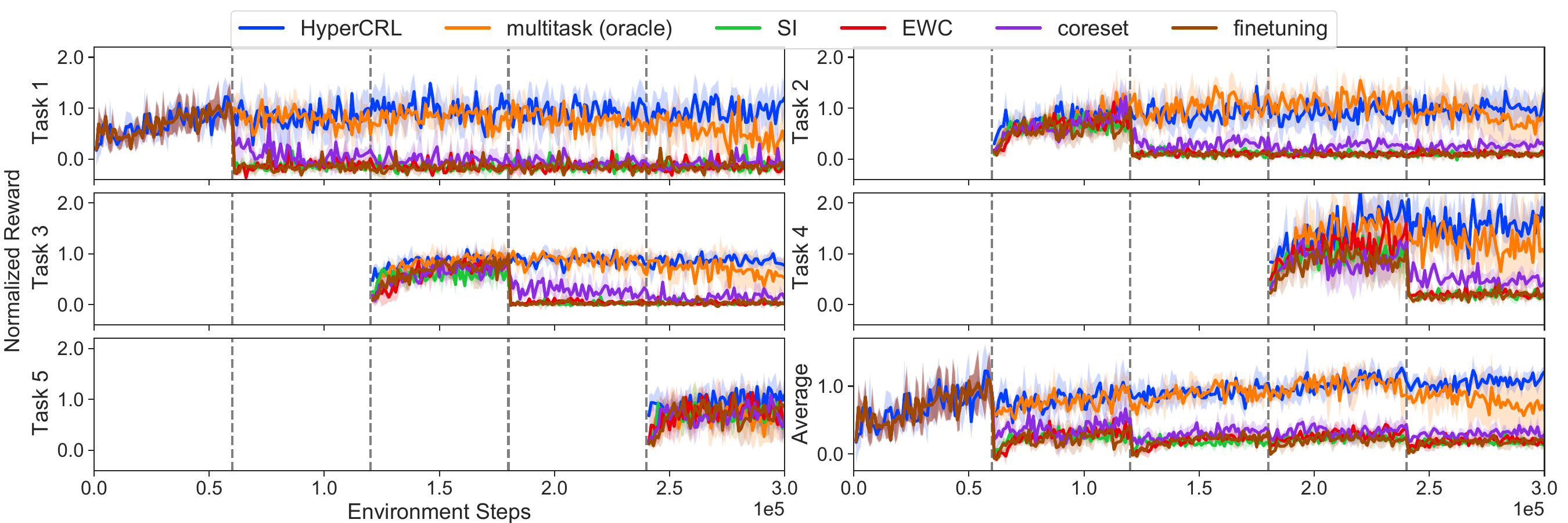}
    \caption{\textbf{Normalized Reward on Door Environment} during training. Results are averaged across four random seeds, and the error bars represent one standard deviation. Each task is trained for 60k steps, totaling 300k steps. The reward is normalized with respect to a model trained from scratch separately on each task.}
    \label{fig:door_reward}
    \vspace{-0.6cm}
\end{figure*}

\vspace{-0.1cm}
\subsection{Results}
\noindent\textbf{Pusher} \algoName~is able to learn to push all 5 of the blocks across the table with minimal forgetting (Table \ref{table:pusher_forgetting}), and even shows signs of positive backward transfer. The average performance of our method is on par with the multi-task learning baseline (Figure \ref{fig:pusher_reward}), which has access to the entire history of data. \algoName~also outperforms other continual learning baselines, either regularization-based (SI, EWC), or replay-based (Coreset). Simple finetuning, however, catastrophically forgets and is unable to perform pushing for all 5 types of block at the end.

\noindent\textbf{Door}
\algoName~outperforms all continual learning baselines on the door opening task, and the multi-task baseline.
Figure \ref{fig:door_reward} shows the learning curves of all our evaluated methods, compared to a single-task baseline trained from scratch on each task. \algoName~virtually sees no performance degradation in terms of reward (Table \ref{table:pusher_forgetting}).

\noindent\textbf{Block Sliding} \algoName~ outperforms all the continual learning baselines although its performance falls short of the multi-task baseline. In Table \ref{table:pusher_forgetting}, \algoName~experiences a slight degree of forgetting on average. The learning curve is omitted here due to space constraint and is available on our project website.

\subsection{Training details}
For all experiments, we model the target network as a multi-layer perceptron (MLP) with two hidden layers (four for door opening) of 200 neurons each and ReLU non-linearity. Each task embedding is initialized as a $10d$ standard normal vector. The hypernetwork is also a MLP with two hidden layers of 50 neurons each (256 neurons for door opening). The parameters of the hypernetwork are initialized with Xavier initialization \cite{glorot2010understanding}. We use Adam with a learning rate of 0.0001 to optimize $\mathcal{L}_t$. During planning, we run CEM for 5 iterations to optimize the actions for a horizon of $h=20$ ($10$ for door opening) steps. Each iteration, we sample 500 (2000 for door opening) action sequences to maximize the sum of rewards. For SI and EWC, we use implementations from \cite{threescenario_cl}.

\section{Limitations and Future Work}
\vspace{-0.05cm}

While our proposed approach shows good results against the baselines we tested, there are many interesting prospects for future work. First, better techniques to train hypernetworks will significantly aid this line of work. Currently, the size of our hypernetwork is at least an order of magnitude larger than the target network, since it directly outputs all the weights. Hypernetworks are often sensitive to the choice of random seeds and architecture. Second, extending our method to image-based RL environments is worth investigating, as it will enable higher-capacity target networks.   
%A challenge could be how to learn meaningful latent dynamics model when there are many modes of ambiguities in observations, for example occlusion. 
Finally, \algoName~is not task agnostic, nor can it automatically detect task switching that happens continuously with no clear task boundaries. A possible direction is to use probabilistic inference models (i.e.~Bayesian non-parametrics \cite{xu2020taskagnostic}) or changepoint detection methods to perform task identification.

\section{Conclusion}
\vspace{-0.05cm}

In this paper we described \algoName, a task-aware method for continual model-based reinforcement learning using hypernetworks. In all of our experiments, we have demonstrated that \algoName~consistently outperforms alternative continual learning baselines in terms of overall performance at the end of training, as well as in terms of retaining performance on previous tasks. %The gap in performance is particularly accentuated in the door opening environment where \algoName~exceeds the performance of the multitask oracle method with access to all previous task data\Florian{We can't explain why this is happening though. Have we tuned the multi-task settings sufficiently?}.\Philip{no not the multi-task setting. The only thing to tune is pretty much make network larger}
By allowing the entire network to change between tasks, rather than just the output head layer of the dynamics network, \algoName~is more effective at accurately representing different dynamics, even with significant task switching, while only requiring a fixed-size hypernetwork and no training data from old tasks to generate task-conditional dynamics for lifelong planning.

%===============================================================================

\bibliographystyle{unsrt}
\bibliography{example}  % .bib
\newpage

\begin{center}
\Large{\textbf{Appendix}}
\end{center}
\section{Additional Test Results}
The objective of a continual learning algorithm is to continuously learn new knowledge on the current task, while maintaining its performance on past tasks. To achieve this, there are two components that needs to be quantified: 1) \textbf{backward transfer} and 2) \textbf{forward transfer}. We formally define them in the context of continual reinforcement learning as follow:

\noindent\textbf{Backward Transfer} First, let $r_{i, j}$ denote the test episodic (normalized) reward on task $i$ after training on task $j$. Retention for each task is then given by $f_i = r_{i, T} / r_{i, i}$ for all $i < T$. The average retention is $f = \frac{1}{T-1} \sum_{i=1}^{T-1} f_i$. If this measure is larger than 100\%, it indicates \textit{positive backward transfer}, which means the performance on previously seen tasks benefit from more recent experiences. If this measure of retention is lower than $100\%$, this means that the robot forgets some knowledge about older tasks and we call this \textit{negative backward transfer} or \textit{forgetting}.

\begin{table} [b!]
\vspace*{-0.2cm}
 \begin{tabular}{ p{1.4cm}p{1.4cm}p{1.4cm}p{1.4cm} p{1.4cm}   p{1.4cm}  }
 \hline
 \multicolumn{6}{c}{ \textbf{\% Forward Transfer in terms of Task Reward (Pusher)}} \\
 \hline
 \textbf{Task} & \textbf{2} & \textbf{3} & \textbf{4} & \textbf{5} & \textbf{Average} \\
 \hline
 Multi-task&  118 $\pm$ 23 & 106 $\pm$ 12 & \textbf{105 $\pm$ 9} & 110 $\pm$ 21 & 109 $\pm$ 9 \\
 HyperCRL &  127 $\pm$ 22 & 99 $\pm$ 12 & 94 $\pm$ 12 & 107 $\pm$ 20 & 107 $\pm$ 9 \\
 SI &  114 $\pm$ 26  & 84 $\pm$ 13  & 98 $\pm$ 15 & 101 $\pm$ 20 & 99 $\pm$ 10 \\
 EWC & 125 $\pm$ 23 & \textbf{109 $\pm$ 11} & 105 $\pm$ 10 & \textbf{110 $\pm$ 21} & \textbf{112 $\pm$ 9} \\
 Coreset & 126 $\pm$ 20  & 99 $\pm$ 13 & 90 $\pm$ 17 & 106 $\pm$ 21 & 105 $\pm$ 9 \\
 Finetuning & \textbf{138 $\pm$ 22} & 108 $\pm$ 9 & 94 $\pm$ 17 & 107 $\pm$ 23 & 112 $\pm$ 9 \\
 \hline
 \hline
 \multicolumn{6}{c}{ \textbf{\% Forward Transfer in terms Task Normalized Reward (Door)}} \\
 \hline
\textbf{Task} & \textbf{2} & \textbf{3} & \textbf{4} & \textbf{5} & \textbf{Average} \\
 \hline
 Multi-task&  103 $\pm$ 74 & \textbf{94 $\pm$ 34} & 122 $\pm$ 160 & 78 $\pm$ 78 & 96 $\pm$ 49 \\
 HyperCRL  &  \textbf{106 $\pm$ 76} & 91 $\pm$ 31 & \textbf{168 $\pm$ 203} & \textbf{102 $\pm$ 67} & \textbf{117 $\pm$ 57} \\
 SI  &  69 $\pm$ 57  & 56 $\pm$ 40  & 107 $\pm$ 137 & 66 $\pm$ 66 & 75 $\pm$ 42 \\
 EWC  & 71 $\pm$ 62  & 78 $\pm$ 32 & 125 $\pm$ 151 & 71 $\pm$ 61 & 86 $\pm$ 44 \\
 Coreset & 94 $\pm$ 80 & 83 $\pm$ 34 & 113 $\pm$ 145 & 57 $\pm$ 68 & 84 $\pm$ 46 \\ 
 Finetuning & 73 $\pm$ 83 & 90 $\pm$ 43 & 101 $\pm$ 145 & 83 $\pm$ 77 & 87 $\pm$ 47 \\ 
 \hline
 \hline
 \multicolumn{6}{c}{ \textbf{\% Forward Transfer in terms Task Reward (Block Sliding)}} \\
 \hline
\textbf{Task} & \textbf{2} & \textbf{3} & \textbf{4} & \textbf{5} & \textbf{Average} \\
 \hline
 Multi-task&  71 $\pm$ 13 & 92 $\pm$ 20 & 99 $\pm$ 9 & 93 $\pm$ 14 & 89 $\pm$ 7 \\
 HyperCRL  &  \textbf{92 $\pm$ 8} & 97 $\pm$ 14 & 82 $\pm$ 39 & \textbf{94 $\pm$ 14} & 91 $\pm$ 11 \\
 SI  &  91 $\pm$ 7  & 83 $\pm$ 48  & 102 $\pm$ 9 & 88 $\pm$ 14 & 91 $\pm$ 13 \\
 EWC  & 90 $\pm$ 9  & \textbf{115 $\pm$ 14} & 103 $\pm$ 11 & 94 $\pm$ 20 & \textbf{100 $\pm$ 7} \\
 Coreset & 65 $\pm$ 22 & 108 $\pm$ 14 & \textbf{110 $\pm$ 16} & 94 $\pm$ 17 & 95 $\pm$ 9 \\ 
 Finetuning & 48 $\pm$ 42 & 88 $\pm$ 19 & 97 $\pm$ 25 & 76 $\pm$ 29 & 77 $\pm$ 15 \\ 
 \hline
\end{tabular}
\caption{\textbf{Forward Transfer on Pusher, Door and Block Sliding Envs}. We measure the average episodic reward at the end of training on task $t$ compared to training a single-task baseline on the same task from scratch. Results show the mean and one standard deviation, evaluated across four seeds and 10 episodes per seed. Greater than 100 indicates positive forward transfer, otherwise it indicates negative forward transfer.}
\label{table:transfer}
\end{table}

\noindent\textbf{Forward Transfer} Forward Transfer measures how the robot adapts to new tasks after training on older task(s). Specifically, let $r_i^*$ denote the test episodic reward on task $i$ for a single-task model trained from scratch using CEM planning. This single-task model serves as a baseline for performance. We then calculate forward transfer as $I_i = r_{i, i} / r_i^*$, and the average is $I = \frac{1}{T-1} \sum_{i=2}^{T} I_i$. If this measure is greater than 100\%, it indicates \textit{positive forward transfer}, which means that the robot benefits from older experiences when learning a new task. If this measure is lower than $100\%$, it means that the robot is unable to take advantage of past experience to solve new tasks, also known as \textit{negative forward transfer}. 

To better illustrate the continual learning capability of \algoName, we provide an evaluation of the positive transfer capabilities defined above. We show the result on all three experiments in Table \ref{table:transfer}.

In the pusher experiment, \algoName~ achieves an average forward transfer of $100 \pm 11\%$. This means that our proposed method can adapt to new tasks and reach similar level of performances compared to a single-task baseline trained from scratch. In the door experiment, \algoName~ beats all baseline in terms of forward transferring capabilities. In the block sliding experiment, \algoName~ achieves slightly lower average forward transfer compared to the best baseline. A note of caution is that some results have very large variance across different random seeds. Better techniques to train hypernetworks or plan with learned dynamics model would improve the reliability of our approach, and we leave this to future work.

\section{Additional Experiment on HalfCheetah}
\begin{figure*} [t!]
    \vspace{-9cm}
    \centering
    {\includegraphics[width=\textwidth]{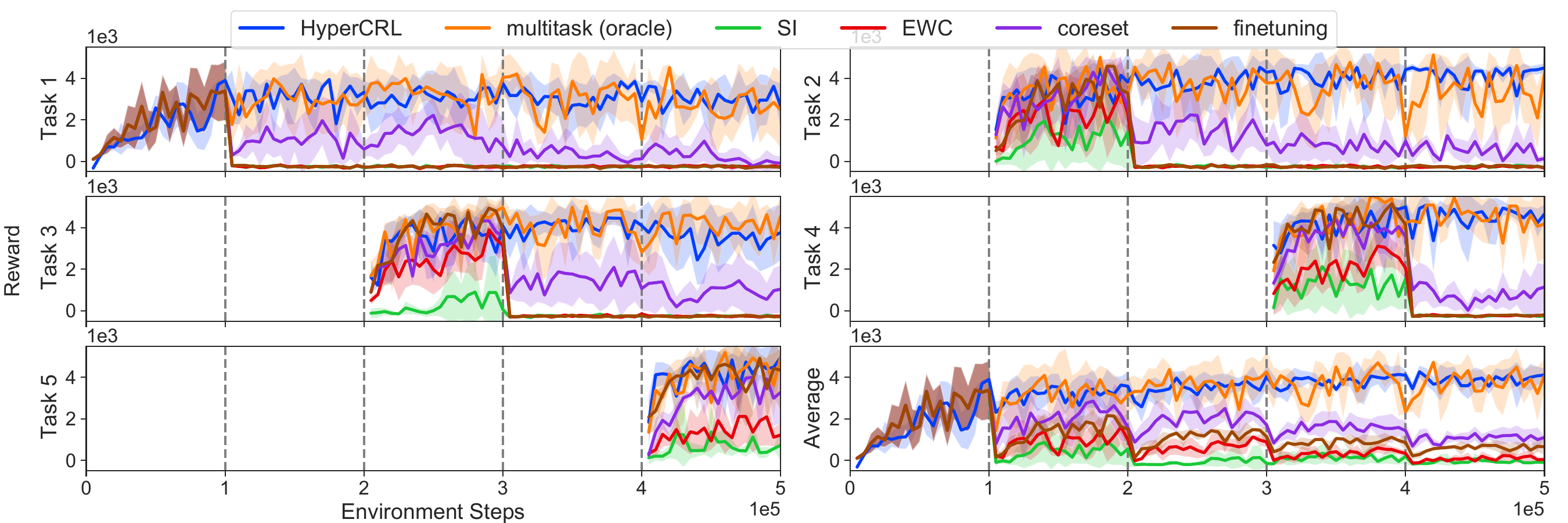}}
    {\caption{\textbf{Reward on HalfCheetah Environment} from \cite{henderson2017benchmark}. Shown are episodic rewards evaluated during training. Each task is trained for 100k steps, totaling 500k steps. Results are averaged across four random seeds,  and shaded area marks one standard deviation.
    \label{fig:cheetah_reward}}}
\end{figure*}

We ran another experiment using \algoName~following the environment setup from \cite{henderson2017benchmark}, changing the body of the robot. We modified body size (torso, leg) on the HalfCheetah environment to create a continual learning scenario with $T=5$ tasks. Our method outperforms other continual learning baselines (SI, EWC, coreset) and performs similarly to multi-task learning (Figure \ref{fig:cheetah_reward}).

\section{More Training Details}
In this section, we provide more details about the hyperparameters used during the experiments in Table \ref{table:hyperparameter}.

\begin{table}[h!]  
     \centering
     \begin{tabular} {p{1.8cm} | p{1cm}| p{1cm} | p{1cm}|p{1cm} | p{1cm}|p{1cm} }
        \hline 
        & $P$ & $M$ & $K$ & $S$ & $\alpha_H$ & $\alpha_\mathbf{e}$  \\
        \hline
        Pusher & 10 & 20 & 200 & 2000 & 0.0001 & 0.0001  \\
        Door & 10 & 300 & 200 & 200 & 0.0001 & 0.0001 \\
        Block Sliding & 10 & 100 & 150 & 500 & 0.0001 & 0.0001 \\
        Half Cheetah & 10 & 100 & 1000 & 2000 & 0.0001 & 0.0001 \\ 
        \hline
        \hline
    \end{tabular}
    \begin{tabular}{p{1.85cm} | p{1cm} | p{1cm} | c | c  }
        & $\mathcal{B}$  & $\beta_\text{reg}$ & hnet non-linearity & hnet hidden layers   \\
        \hline 
        Pusher & 100 & 0.05 & ELU & [50, 50]\\
        Door & 100 & 0.5 & ReLU & [256, 256]\\
        Block Sliding & 100 & 0.5 & ReLU & [50, 50]\\
        Half Cheetah & 100 & 0.05 & ReLU & [256, 256]  \\ 
        \hline 
        \end{tabular}
    
    \caption{Hyper-parameter values of the proposed algorithm~\algoName~ on pusher pusher, door, block sliding and half cheetah environments.}
    \label{table:hyperparameter}
\end{table}

\section{Additional ``Oracle" baseline}
\begin{figure*} [t!]
    \vspace{-18cm}
    \centering
    {\includegraphics[width=\textwidth]{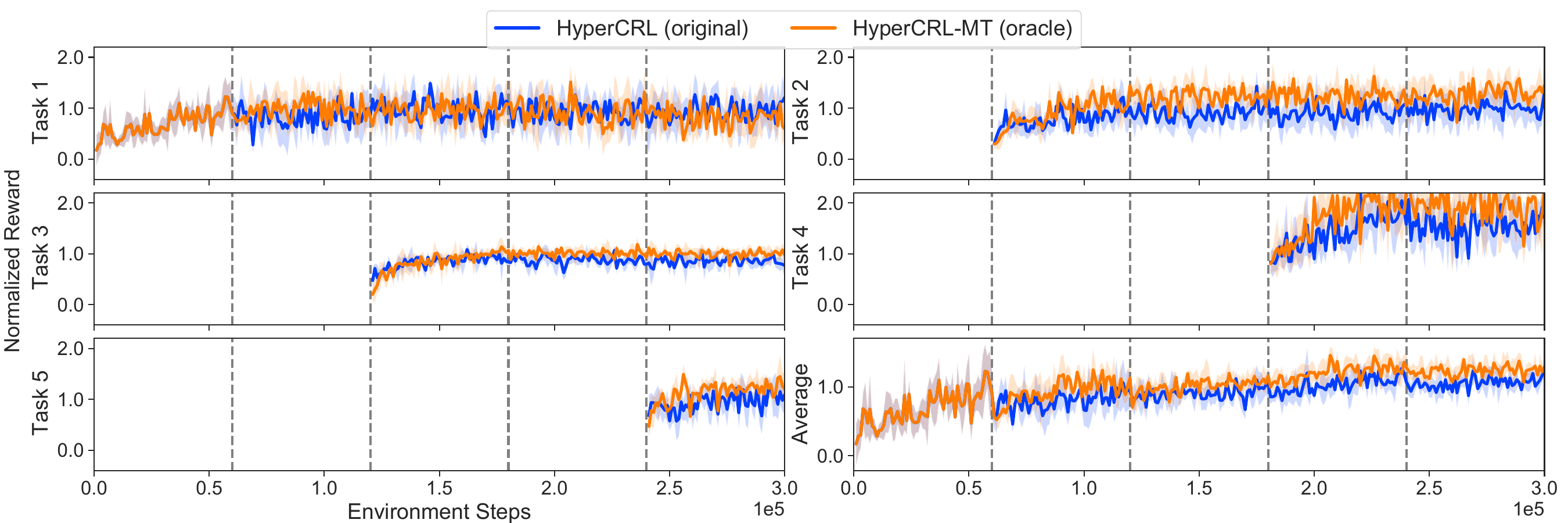}}
    \caption{\textbf{Additional Hypernetwork Baseline on Door Environment} Shown are normalized episodic rewards evaluated during training. ~\algoName\texttt{-MT} is trained with access to the entire replay buffer. Results are averaged across four random seeds, and normalized with respect to a model trained from scratch separately on each task.}
    \label{fig:door_reward_supp_2}
\end{figure*}

In Figure \ref{fig:door_reward}, the multi-head multi-task baseline slightly underperforms \algoName. We hypothesize that the hypernetwork architecture might be more effective for learning the task.
We consider another multi-task baseline that shares the same architecture of the hypernetwork used in \algoName, but is trained without regularization and has access to the entire replay buffer similar to the multi-task baseline. We call this \algoName\texttt{-MT}, and we believe this should serve as a better upper-bound on the continual learning performance of \algoName~. We compare results on  the door-opening experiment in Figure \ref{fig:door_reward_supp_2}. With full access to all past experiences, \algoName\texttt{-MT} slightly outperforms \algoName~ on this experiment, which shows the efficacy of our proposed method.

\end{document}